\documentclass{article}

\usepackage{PRIMEarxiv}

\usepackage[utf8]{inputenc}
\usepackage[T1]{fontenc}
\usepackage{hyperref}
\usepackage{url}
\usepackage{booktabs}
\usepackage{amsfonts}
\usepackage{nicefrac}
\usepackage{microtype}
\usepackage{graphicx}
\usepackage{multirow}
\usepackage{array}
\usepackage{siunitx}
\usepackage{xcolor}
\usepackage{listings}
\usepackage{longtable}
\usepackage{rotating} 

\usepackage[noabbrev,nameinlink]{cleveref}

\definecolor{codegreen}{rgb}{0,0.6,0}
\definecolor{codegray}{rgb}{0.5,0.5,0.5}
\definecolor{codepurple}{rgb}{0.58,0,0.82}
\definecolor{backcolour}{rgb}{0.97,0.97,0.97}

\title{\includegraphics[width=0.99\textwidth]{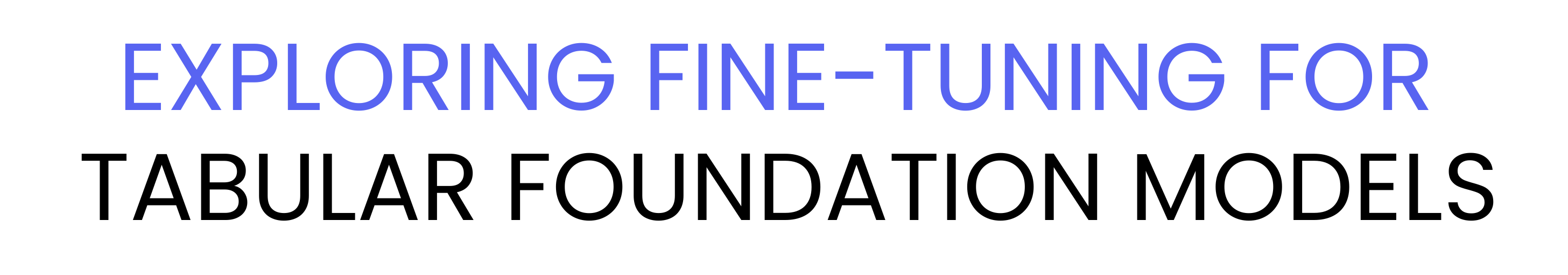}}

\author{
  Aditya Tanna, Pratinav Seth, Mohamed Bouadi, Vinay Kumar Sankarapu \\
  \affiliation{Lexsi Labs, India \& France} \\
  \{aditya.tanna,v.k\}@lexsi.ai \\
}
\setabstract{%
Tabular Foundation Models (TFMs) have recently shown strong in-context learning capabilities on structured data, achieving zero-shot performance comparable to traditional machine learning methods. We find that zero-shot TFMs already achieve strong performance, while the benefits of fine-tuning are highly model- and data-dependent. Meta-learning and PEFT provide moderate gains under specific conditions, whereas full supervised fine-tuning (SFT) often reduces accuracy or calibration quality.
This work presents the first comprehensive study of fine-tuning in TFMs across benchmarks including TALENT, OpenML-CC18, and TabZilla. We compare Zero-Shot, Meta-Learning, Supervised (SFT), and parameter-efficient (PEFT) approaches, analyzing how dataset factors—such as imbalance, size, and dimensionality—affect outcomes. Our findings cover performance, calibration, and fairness, offering practical guidelines on when fine-tuning is most beneficial and its limitations.
}
\setkeywords{Tabular Foundation Models, Meta-Learning, PEFT, Calibration, Fairness, Benchmarking.}
\runningtitle{Exploring Fine-Tuning for Tabular Foundation Models}

\begin{document}
\maketitle
\section{Introduction}
Tabular data drives applications in enterprises like healthcare, finance, and analytics. For decades, gradient-boosted decision trees (GBDTs) such as XGBoost~\cite{chen2016xgboost}, LightGBM~\cite{ke2017lightgbm}, and CatBoost~\cite{prokhorenkova2018catboost} have dominated due to their robustness on heterogeneous features and small datasets. While deep learning has revolutionized vision and NLP, standard neural networks often struggle to outperform GBDTs on tabular tasks~\cite{borisov2022tabular,somvanshi2024tabular}, though recent architectures like FT-Transformer~\cite{fttransformer} and TabR~\cite{tabr} have narrowed this gap.

Tabular Foundation Models (TFMs) offer a new paradigm, leveraging large-scale pretraining to enable in-context learning (ICL) and zero-shot generalization. \textsc{TabPFN}~\cite{tabpfn,TabPFN2} approximates Bayesian inference using transformers trained on synthetic data, effectively performing posterior inference in a single forward pass for small datasets. Building on this, \textsc{TabICL}~\cite{tabicl} scales ICL to larger datasets via a two-stage column-then-row attention mechanism. To address high-dimensional dependencies, \textsc{OrionMSP}~\cite{orionmsp2025} and \textsc{OrionBiX}~\cite{orionbix2025} introduce multi-scale sparse attention and biaxial attention mechanisms, respectively, enabling efficient context modeling. \textsc{TabDPT}~\cite{tabdpt} employs diffusion-based pretraining to learn robust representations, while \textsc{Mitra}~\cite{mitra} utilizes mixed synthetic priors to improve generalization.

While zero-shot inference is powerful, adapting TFMs to target distributions can enhance performance. \textsc{Meta-learning} (episodic fine-tuning) preserves ICL capabilities by training on support--query splits, mimicking the inference-time environment~\cite{finn2017model}. \textsc{Supervised fine-tuning (SFT)} updates all parameters on labeled data~\cite{howard2018universal}, which can yield gains on large datasets but risks overfitting on small ones. 
\textsc{Parameter-efficient fine-tuning (PEFT)}, particularly Low-Rank Adaptation (LoRA)~\cite{hu2021lora}, offers a resource-efficient alternative by updating only low-dimensional subspaces, often matching full fine-tuning performance with lower overhead~\cite{liu2024peft,razuvayevskaya2024peft}. 
Strategies like \textsc{LoCalPFN} (Thomas et al.~\cite{localpfn}) have also emerged, combining retrieval with fine-tuning to adapt to local data sub-manifolds.

Beyond predictive performance, responsible deployment requires assessing model trustworthiness through calibration and fairness. Calibration ensures that predicted probabilities align with true outcomes, a critical requirement for high-stakes decision-making in healthcare and finance. Fairness necessitates that models avoid systematic bias across demographic groups. While TFMs show promise in zero-shot settings, the impact of different adaptation strategies on these safety-critical dimensions remains under-explored.
However, it remains unclear \emph{when} fine-tuning helps these models, \emph{which} strategies are effective, and \emph{how} dataset properties influence outcomes. The wide variation in tabular dataset size, imbalance, and feature structure makes the impact of fine-tuning particularly non-obvious. This work presents the first large-scale study of fine-tuning TFMs. 

\begin{figure}[pt]
    \centering
    \includegraphics[width=0.99\linewidth]{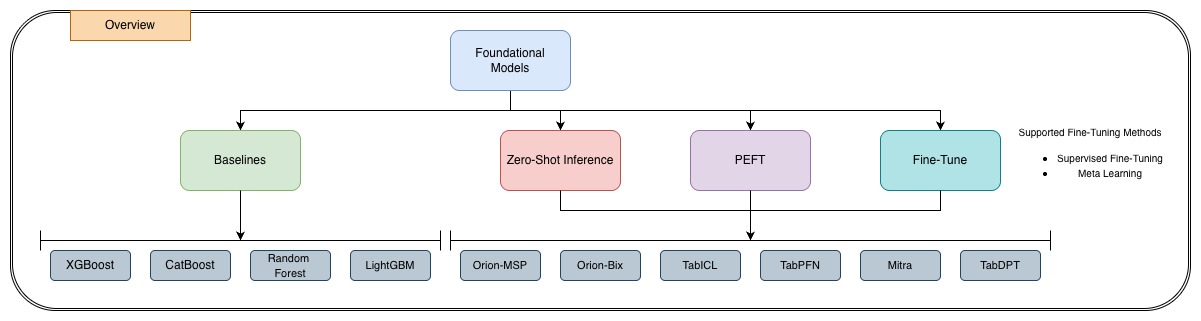}
    \caption{High-level overview of models and adaptation strategies. We benchmark classical tabular baselines and multiple tabular foundation models under three regimes—zero-shot inference, parameter-efficient fine-tuning (PEFT), and full supervised fine-tuning—using both supervised fine-tuning and meta-learning variants}
    \label{fig:finetune}
\end{figure}

Our contributions are: (I) a unified evaluation of four adaptation strategies—zero-shot, meta-learning, supervised fine-tuning (SFT), and parameter-efficient fine-tuning (PEFT)—across six TFMs and multiple benchmarks; (II) an analysis of how dataset factors (size, imbalance, dimensionality) shape fine-tuning outcomes; (III) an assessment of calibration and fairness effects; and (IV) practical guidelines on when fine-tuning is effective. Our findings show that fine-tuning is not universally beneficial, underscoring the need for careful strategy selection. An Overview of all the models and adaptation strategies in shown in Figure \ref{fig:finetune}

\section{Methodology}
\label{sec:methodology}

\subsection{Models and Benchmarks}
We evaluate six Tabular Foundation Models—TabPFN, TabICL, OrionMSP, OrionBiX, TabDPT, and Mitra—alongside classical baselines (XGBoost, LightGBM, CatBoost, Random Forest). Experiments span three benchmark suites: \textsc{TALENT} (155 datasets)~\cite{talent}, \textsc{OpenML-CC18} (63 datasets)~\cite{openmlcc18}, and \textsc{TabZilla} (27 datasets)~\cite{tabzilla}. For fairness evaluation, we additionally use nine public datasets with sensitive attributes, including \emph{Adult Census Income}, \emph{German Credit}, and \emph{COMPAS Recidivism}. All experiments follow standardized splits and are executed on NVIDIA L40S and H200 GPUs. We utilize TabTune~\cite{tanna2025tabtune} for TFM experiments and it's AutoGluon~\cite{erickson2020autogluon} implementation for baselines.

\subsection{Adaptation Strategies}
We compare four distinct adaptation paradigms to assess their impact on performance, calibration, and fairness:
\begin{itemize}
    \item \textbf{Zero-Shot Inference}: Evaluates pretrained TFMs directly on target tasks without parameter updates, relying solely on in-context learning or prior knowledge.
    \item \textbf{Meta-Learning Fine-Tuning}: Employ episodic training (48--512 support samples per episode for 5 epochs) to adapt the model while preserving its in-context learning capabilities.
    \item \textbf{Supervised Fine-Tuning (SFT)}: Performs full-parameter optimization using the AdamW optimizer with learning rates between $1\text{e}^{-5}$ and $5\text{e}^{-5}$, employing early stopping (up to 10 epochs) to mitigate overfitting.
    \item \textbf{Parameter-Efficient Fine-Tuning (PEFT)}: Utilizes Low-Rank Adaptation (LoRA)~\cite{hu2021lora} with rank $r{=}8$ and $\alpha{=}16$. This strategy is applied to both supervised and meta-learning regimes to reduce memory overhead while maintaining adaptation flexibility.
\end{itemize}

\begin{table}[pt]
  \scriptsize
  \centering
  \caption{Overall leaderboard across three benchmark suites—TALENT, OpenML-CC18, and TabZilla. Ranks denote the mean rank per benchmark suite (lower is better). Metrics: ACC = Accuracy, F1 = Weighted F1. The "All" column reports the aggregated rank across for a strategy. Models are grouped by adaptation strategy. Formatting: \textbf{\underline{1st place}}; \underline{2nd place} within each group. \textit{Note : that rank reflects only relative ordering and may not capture absolute ACC/F1 trends, which can vary across scenarios.}}
  \vspace{0.12in}
  \label{tab:overall}
  \begin{tabular}{l c ccc ccc ccc}
    \toprule
Strategy
  & \multicolumn{1}{c}{All}
  & \multicolumn{3}{c}{TALENT}
  & \multicolumn{3}{c}{OpenML-CC18}
  & \multicolumn{3}{c}{TabZilla} \\
\cmidrule(lr){1-1} \cmidrule(lr){2-2} \cmidrule(lr){3-5} \cmidrule(lr){6-8} \cmidrule(lr){9-11}
Models & Rank($\downarrow$) & Rank($\downarrow$) & ACC($\uparrow$) & F1($\uparrow$) & Rank($\downarrow$) & ACC($\uparrow$) & F1($\uparrow$) & Rank($\downarrow$) & ACC($\uparrow$) & F1($\uparrow$) \\
    \midrule
\multicolumn{11}{l}{\textit{Baselines + Zero-Shot Inference}} \\
\cmidrule(lr){1-11}
    XGBoost    & 6.39 & 5.97 & 0.8403 & 0.8360 & 5.34 & 0.8558 & 0.8537 & 5.62 & 0.8612 & 0.8326 \\
    CatBoost  & 6.16 & 5.54 & 0.8336 & 0.8259 & 6.64 & 0.8588 & 0.8520 & 5.40 & 0.8579 & 0.8384 \\
    Random Forest & 7.06 & 6.11 & 0.8285 & 0.8209 & 5.78 & 0.8547 & 0.8497 & 7.72 & 0.8358 & 0.8399 \\
    LightGBM  & 6.50 & 6.07 & 0.8331 & 0.8245 & 5.60 & 0.8581 & 0.8493 & 4.90 & 0.8618 & 0.8211 \\
    TabICL & 4.78 & 4.07 & \underline{0.8471} & \underline{0.8379} & 4.36 & 0.8667 & 0.8623 & 5.63 & 0.8734 & 0.8698 \\
    OrionBiX & 5.27 & 4.56 & 0.8346 & 0.8260 & 4.97 & 0.8653 & 0.8596 & 4.63 & 0.8728 & 0.8628 \\
    \textbf{\underline{OrionMSP}} & \textbf{\underline{3.76}} & \textbf{\underline{3.26}} & 0.8461 & 0.8360 & \textbf{\underline{3.90}} & \textbf{\underline{0.8722}} & \textbf{\underline{0.8676}} & \textbf{\underline{3.78}} & \textbf{\underline{0.8821}} & \textbf{\underline{0.8786}} \\
    \underline{TabPFN} & \underline{4.44} & \underline{3.71} & \textbf{\underline{0.8514}} & \textbf{\underline{0.8412}} & 4.36 & \underline{0.8714} & \underline{0.8663} & 4.65 & 0.8752 & 0.8716 \\
    Mitra  & 10.85 & 9.67 & 0.3921 & 0.2868 & 9.55 & 0.3614 & 0.2522 & 10.21 & 0.3152 & 0.1830 \\
    TabDPT & 5.27 & 5.16 & 0.8408 & 0.8318 & \underline{4.32} & 0.8672 & 0.8625 & \underline{3.81} & \underline{0.8814} & \underline{0.8775} \\
\midrule
\multicolumn{11}{l}{\textit{FineTune – Meta Learning}} \\
\cmidrule(lr){1-11}
    TabICL        & 3.66 & 3.33 & 0.8344 & 0.8253 & 3.05 & 0.8664 & 0.8597 & 4.15 & 0.6956 & 0.6845 \\
    OrionBiX & 3.96 & 4.25 & 0.8158 & 0.8060 & 2.86 & \underline{0.8548} & \underline{0.8516} & \underline{2.00} & \underline{0.8726} & \underline{0.8662} \\
    \textbf{\underline{OrionMSP}}     & \textbf{\underline{2.26}} & \textbf{\underline{1.80}} & \underline{0.8401} & \underline{0.8310} & \underline{2.82} & \underline{0.8548} & \underline{0.8516} & \textbf{\underline{1.73}} & \textbf{\underline{0.8735}} & \textbf{\underline{0.8672}} \\
    \underline{TabPFN}        & \underline{2.42} & \underline{2.07} & \textbf{\underline{0.8517}} & \textbf{\underline{0.8414}} & \textbf{\underline{2.42}} & \textbf{\underline{0.8842}} & \textbf{\underline{0.8784}} & 2.42 & 0.8663 & 0.8603 \\
    Mitra         & 6.09 & 5.79 & 0.6416 & 0.5874 & 5.42 & 0.6164 & 0.5651 & 4.70 & 0.5592 & 0.5147 \\
    TabDPT        & 3.95 & 3.72 & 0.8255 & 0.8167 & 4.40 & 0.8534 & 0.8501 & 2.62 & 0.8500 & 0.8456 \\
\midrule
\multicolumn{11}{l}{\textit{FineTune – Supervised Fine Tuning}} \\
\cmidrule(lr){1-11}
    TabICL        & 5.05 & 4.74 & 0.7668 & 0.7439 & 4.36 & 0.6838 & 0.6299 & 4.65 & 0.5670 & 0.4733 \\
    OrionBiX & 4.21 & 3.54 & 0.7698 & 0.7469 & 4.30 & 0.7126 & 0.6595 & 4.25 & 0.6476 & 0.5782 \\
    OrionMSP     & 2.88 & 2.27 & 0.7908 & 0.7653 & 3.18 & 0.7995 & 0.7668 & 2.88 & 0.7454 & 0.7222 \\
    \textbf{\underline{TabPFN}}        & \underline{\textbf{1.97}} & \textbf{\underline{1.83}} & \textbf{\underline{0.8459}} & \textbf{\underline{0.8350}} & \textbf{\underline{1.89}} & \textbf{\underline{0.8697}} & \textbf{\underline{0.8617}} & 1.86 & \textbf{\underline{0.8433}} & \textbf{\underline{0.8327}} \\
    Mitra         & 5.98 & 5.62 & 0.5460 & 0.4382 & 5.02 & 0.5408 & 0.4309 & 5.52 & 0.4608 & 0.3467 \\
    \underline{TabDPT}        & \underline{2.79} & 2.98 & 0.8202 & 0.8094 & \underline{2.25} & \underline{0.8499} & \underline{0.8424} & \textbf{\underline{1.81}} & \underline{0.8337} & \underline{0.8260} \\
\midrule
\multicolumn{11}{l}{\textit{FineTune – PEFT - Meta Learning}} \\
\cmidrule(lr){1-11}
    TabICL        & 4.32 & 4.07 & 0.7017 & 0.6867 & 4.07 & 0.7773 & 0.7605 & 3.45 & 0.7116 & 0.7003 \\
    OrionBiX     & 2.77 & 2.39 & 0.7854 & \underline{0.7762} & 2.50 & 0.8471 & 0.8430 & 3.52 & 0.7370 & 0.7200 \\
    \textbf{\underline{OrionMSP}}     & \underline{\textbf{2.21}} & \underline{2.25} & \underline{0.7879} & 0.7728 & \textbf{\underline{1.93}} &\underline{ 0.8566} & 0.8453 & \textbf{\underline{1.33}} & \textbf{\underline{0.8594}} & \textbf{\underline{0.8581}} \\
    Mitra         & 4.64 & 4.17 & 0.6369 & 0.5905 & 4.32 & 0.6000 & 0.5426 & 4.36 & 0.5461 & 0.4960 \\
    \underline{TabDPT}        & \underline{2.28} & \textbf{\underline{2.11}} & \textbf{\underline{0.8002}} & \textbf{\underline{0.7910}} & \underline{2.17} & \textbf{\underline{0.8600}} & \textbf{\underline{0.8539}} & 2.33 & \underline{0.8495} & \underline{0.8481} \\
\midrule
\multicolumn{11}{l}{\textit{FineTune – PEFT - Supervised Fine Tuning}} \\
\cmidrule(lr){1-11}
    TabICL        & 3.70 & 3.72 & 0.5692 & 0.4638 & 3.06 & 0.8174 & 0.7965 & 2.92 & 0.7920 & 0.7776 \\
    OrionBiX     & 3.99 & 3.57 & 0.6358 & 0.5570 & 3.67 & 0.7380 & 0.6789 & 3.95 & 0.6707 & 0.6071 \\
    \underline{OrionMSP}     & \underline{2.01} & \underline{2.01} & \underline{0.6749} & 0.5956 & \textbf{\underline{1.73}} & \underline{0.8241} & \underline{0.8033} & \textbf{\underline{1.47}} & \underline{0.8214} & \underline{0.8071} \\
    Mitra         & 4.62 & 4.01 & 0.5294 & 0.4303 & 4.49 & 0.4965 & 0.3917 & 4.70 & 0.4606 & 0.3597 \\
    \textbf{\underline{TabDPT}}        & \textbf{\underline{1.91}} & \textbf{\underline{1.65}} & \textbf{\underline{0.7988}} & \textbf{\underline{0.7842}} & \underline{2.03} & \textbf{\underline{0.8500}} & \textbf{\underline{0.8398}} & \underline{1.95} & \textbf{\underline{0.8461}} & \textbf{\underline{0.8461}} \\
\bottomrule
\end{tabular}
\end{table}

\newpage
\subsection{Evaluation Metrics}
Our assessment covers three complementary dimensions:
\begin{itemize}
    \item \textbf{Performance}: Measured via Accuracy, weighted F1-score, and mean rank across datasets.
    \item \textbf{Calibration}: Assessed using Expected Calibration Error (ECE), Maximum Calibration Error (MCE), and Brier Score~\cite{guo2017calibration} to quantify the reliability of probability estimates.
    \item \textbf{Fairness}: Evaluated using Statistical Parity Difference (SPD), Equalized Odds Difference (EOD), and Equal Opportunity Difference (EOpD)~\cite{barocas2019fairness,hardt2016equality,pleiss2017fairness} to detect group disparities.
\end{itemize}

\subsection{Research Questions}
In this work, we study: (1) whether fine-tuning improves TFM performance compared to zero-shot inference; (2) how dataset properties such as size, imbalance, and dimensionality influence these gains; (3) how different adaptation strategies affect model calibration; and (4) the implications of fine-tuning on fairness.

\section{Experimental Results}
\label{sec:experiments}

\begin{sidewaystable}[pt]
  \scriptsize
  \centering
  \caption{Comprehensive performance analysis across dataset characteristics: size (Small $<$1K, Medium 1K--10K, Large $>$10K samples), class balance (Balanced $\ge$0.6, Imbalanced $<$0.6), and feature dimensionality (Narrow $<$10, Medium 10--100, Wide $>$100 features). ACC = Accuracy; F1 = Weighted F1-score. Values are averaged across datasets within each category. Models are grouped by adaptation strategy. Formatting: \textbf{\underline{1st place}}; \underline{2nd place} within each group.}
  \vspace{0.12in}
  \label{tab:comprehensive-analysis}
  \begin{tabular}{l cc cc cc cc cc cc cc cc cc cc cc}
    \toprule
    & \multicolumn{6}{c}{Dataset Size} & \multicolumn{4}{c}{Dataset Imbalance} & \multicolumn{6}{c}{Dataset Width} \\
    \cmidrule(lr){2-7} \cmidrule(lr){8-11} \cmidrule(lr){12-17}
Strategy & \multicolumn{2}{c}{Small (<1K)} & \multicolumn{2}{c}{Medium (1K-10K)} & \multicolumn{2}{c}{Large (>10K)} & \multicolumn{2}{c}{Balanced ($\ge$0.6)} & \multicolumn{2}{c}{Imbalanced (<0.6)} & \multicolumn{2}{c}{Narrow (<10)} & \multicolumn{2}{c}{Medium (10-100)} & \multicolumn{2}{c}{Wide (>100)} \\
\cmidrule(lr){1-1} \cmidrule(lr){2-3} \cmidrule(lr){4-5} \cmidrule(lr){6-7} \cmidrule(lr){8-9} \cmidrule(lr){10-11} \cmidrule(lr){12-13} \cmidrule(lr){14-15} \cmidrule(lr){16-17}
Models & ACC($\uparrow$) & F1($\uparrow$) & ACC($\uparrow$) & F1($\uparrow$) & ACC($\uparrow$) & F1($\uparrow$) & ACC($\uparrow$) & F1($\uparrow$) & ACC($\uparrow$) & F1($\uparrow$) & ACC($\uparrow$) & F1($\uparrow$) & ACC($\uparrow$) & F1($\uparrow$) & ACC($\uparrow$) & F1($\uparrow$) \\
    \midrule
\multicolumn{17}{l}{\textit{Baselines + Zero-Shot Inference}} \\
\cmidrule(lr){1-17}
    XGBoost & 0.8168 & 0.7964 & 0.8363 & 0.8314 & \textbf{\underline{0.8969}} & \textbf{\underline{0.8920}} & 0.8175 & 0.8110 & \textbf{\underline{0.8859}} & \textbf{\underline{0.8785}} & 0.8222 & 0.8159 & 0.8482 & 0.8410 & 0.9140 & 0.9039 \\
    CatBoost & 0.8124 & 0.7935 & 0.8340 & 0.8264 & 0.8797 & 0.8733 & 0.8076 & 0.8020 & 0.8785 & 0.8665 & 0.8145 & 0.8067 & 0.8441 & 0.8344 & \textbf{\underline{0.9157}} & \underline{0.9084} \\
    Random Forest & 0.7988 & 0.8187 & 0.8285 & 0.8221 & 0.8694 & 0.8628 & 0.7983 & 0.7955 & 0.8741 & 0.8646 & 0.8005 & 0.7044 & 0.8410 & 0.8235 & 0.9034 & 0.8936 \\
    LightGBM & 0.8143 & 0.7789 & 0.8314 & 0.8226 & 0.8827 & 0.8764 & 0.8071 & 0.7977 & 0.8775 & 0.8633 & 0.8128 & 0.7907 & 0.8458 & 0.8326 & 0.8999 & 0.8908 \\
    TabICL & 0.8301 & \textbf{\underline{0.8338}} & 0.8486 & 0.8398 & 0.8802 & 0.8743 & \underline{0.8279} & \underline{0.8233} & 0.8806 & 0.8698 & 0.8208 & 0.8119 & \underline{0.8627} & \underline{0.8549} & 0.9101 & 0.8936 \\
    OrionBiX & \underline{0.8330} & 0.8150 & 0.8348 & 0.8260 & 0.8729 & 0.8670 & 0.8096 & 0.8040 & 0.8787 & 0.8683 & 0.8112 & 0.8043 & 0.8510 & 0.8417 & 0.8859 & \underline{0.8849} \\
    \underline{OrionMSP} & 0.8232 & 0.8194 & \underline{0.8494} & \underline{0.8402} & \underline{0.8843} & \underline{0.8768} & 0.8265 & 0.8202 & \underline{0.8840} & \underline{0.8731} & \textbf{\underline{0.8394}} & \textbf{\underline{0.8314}} & 0.8572 & 0.8478 & 0.8860 & 0.8837 \\
    TabPFN & 0.8325 & 0.8131 & \textbf{\underline{0.8557}} & \textbf{\underline{0.8462}} & 0.8783 & 0.8713 & \textbf{\underline{0.8367}} & \textbf{\underline{0.8309}} & 0.8808 & 0.8697 & 0.8187 & 0.8092 & \textbf{\underline{0.8676}} & \textbf{\underline{0.8589}} & \underline{0.9129} & \textbf{\underline{0.9111}} \\
    Mitra & 0.4334 & 0.3236 & 0.3600 & 0.2553 & 0.3837 & 0.2754 & 0.2763 & 0.1540 & 0.4794 & 0.3858 & 0.3737 & 0.2683 & 0.3886 & 0.2781 & 0.2521 & 0.1497 \\
    \textbf{\underline{TabDPT}} & \textbf{\underline{0.8333}} & \underline{0.8271} & 0.8424 & 0.8339 & 0.8831 & 0.8765 & 0.8233 & 0.8189 & 0.8798 & 0.8690 & \underline{0.8262} & \underline{0.8189} & 0.8566 & 0.8483 & 0.8845 & 0.8820 \\
\midrule
\multicolumn{17}{l}{\textit{FineTune – Meta Learning}} \\
\cmidrule(lr){1-17}
    TabICL & 0.7912 & 0.7806 & 0.8342 & 0.8256 & 0.8431 & 0.8360 & 0.7943 & 0.7884 & 0.8662 & 0.8547 & 0.7867 & 0.7779 & 0.8383 & 0.8338 & 0.8319 & 0.8667 \\
    OrionBiX & 0.8149 & 0.8123 & 0.8283 & 0.8204 & 0.8410 & 0.8306 & 0.7947 & 0.7881 & 0.8622 & 0.8532 & 0.7956 & 0.7900 & 0.8316 & 0.8278 & 0.8429 & 0.8640 \\
    \underline{OrionMSP} & 0.8211 & 0.8151 & \underline{0.8413} & \underline{0.8330} & \textbf{\underline{0.8787}} & \textbf{\underline{0.8721}} & \underline{0.8231} & \underline{0.8183} & \underline{0.8735} & \underline{0.8636} & \underline{0.8253} & \textbf{\underline{0.8190}} & \underline{0.8514} & \underline{0.8457} & \underline{0.8564} & \underline{0.8737} \\
    \textbf{\underline{TabPFN}} & \textbf{\underline{0.8336}} & \underline{0.8256} & \textbf{\underline{0.8638}} & \textbf{\underline{0.8548}} & \underline{0.8748} & \underline{0.8668} & \textbf{\underline{0.8461}} & \textbf{\underline{0.8407}} & \textbf{\underline{0.8784}} & \textbf{\underline{0.8664}} & \textbf{\underline{0.8264}} & \underline{0.8171} & \textbf{\underline{0.8683}} & \textbf{\underline{0.8579}} & \textbf{\underline{0.9172}} & \textbf{\underline{0.9533}} \\
    Mitra & 0.6417 & 0.5988 & 0.6083 & 0.5516 & 0.6585 & 0.6132 & 0.4699 & 0.4142 & 0.7885 & 0.7391 & 0.6646 & 0.6247 & 0.6004 & 0.5763 & 0.4090 & 0.3224 \\
    TabDPT & \underline{0.8224} & \textbf{\underline{0.8188}} & 0.8289 & 0.8212 & 0.8682 & 0.8613 & 0.8066 & 0.8023 & 0.8650 & 0.8552 & 0.8129 & 0.8071 & 0.8405 & 0.8350 & 0.8400 & 0.8497 \\
\midrule
\multicolumn{17}{l}{\textit{FineTune - Supervised FineTuning}} \\
\cmidrule(lr){1-17}
    TabICL & 0.6746 & 0.6225 & 0.7553 & 0.7187 & 0.6693 & 0.6212 & 0.6708 & 0.6317 & 0.7794 & 0.7362 & 0.6609 & 0.6195 & 0.7507 & 0.7101 & 0.7757 & 0.7269 \\
    OrionBiX & 0.7158 & 0.6792 & 0.7718 & 0.7386 & 0.6697 & 0.6272 & 0.6917 & 0.6550 & 0.7934 & 0.7590 & 0.6922 & 0.6576 & 0.7643 & 0.7294 & 0.7560 & 0.7046 \\
    OrionMSP & 0.8135 & 0.7920 & 0.8036 & 0.7798 & 0.7231 & 0.6828 & 0.7467 & 0.7287 & 0.8307 & 0.7953 & 0.7540 & 0.7297 & 0.8025 & 0.7750 & 0.8148 & 0.7808 \\
    \underline{TabPFN} & \underline{0.8216} & \underline{0.8109} & \textbf{\underline{0.8580}} & \textbf{\underline{0.8485}} & \textbf{\underline{0.8572}} & \textbf{\underline{0.8459}} & \textbf{\underline{0.8336}} & \textbf{\underline{0.8267}} & \textbf{\underline{0.8709}} & \textbf{\underline{0.8578}} & \textbf{\underline{0.8258}} & \textbf{\underline{0.8162}} & \textbf{\underline{0.8578}} & \textbf{\underline{0.8469}} & \textbf{\underline{0.9346}} & \textbf{\underline{0.9335}} \\
    Mitra & 0.5625 & 0.4408 & 0.5133 & 0.4116 & 0.5720 & 0.4501 & 0.3216 & 0.2037 & 0.7413 & 0.6415 & 0.5619 & 0.4478 & 0.5349 & 0.4268 & 0.3307 & 0.2436 \\
    \textbf{\underline{TabDPT}} & \textbf{\underline{0.8227}} & \textbf{\underline{0.8153}} & \underline{0.8377} & \underline{0.8284} & \underline{0.8120} & \underline{0.8002} & \underline{0.8025} & \underline{0.7949} & \underline{0.8578} & \underline{0.8464} & \underline{0.8078} & \underline{0.7948} & \underline{0.8378} & \underline{0.8294} & \underline{0.8623} & \underline{0.8587} \\
\midrule
\multicolumn{17}{l}{\textit{FineTune – PEFT - Meta Learning}} \\
\cmidrule(lr){1-17}
    TabICL & 0.7518 & 0.7397 & 0.7251 & 0.7114 & 0.6962 & 0.6799 & 0.6530 & 0.6460 & 0.8026 & 0.7800 & 0.6375 & 0.6375 & 0.7601 & 0.7444 & 0.7714 & 0.7622 \\
    OrionBiX & 0.8093 & 0.8042 & 0.7995 & 0.7939 & 0.7848 & 0.7665 & 0.7504 & 0.7436 & 0.8528 & 0.8420 & 0.7584 & 0.7584 & 0.8147 & 0.8057 & 0.8251 & 0.8243 \\
    \textbf{\underline{OrionMSP}} & \textbf{\underline{0.8275}} & \textbf{\underline{0.8213}} & \textbf{\underline{0.8134}} & \underline{0.8019} & \underline{0.8013} & \underline{0.7817} & \underline{0.7714} & \underline{0.7623} & \textbf{\underline{0.8608}} & \underline{0.8433} & \underline{0.7836} & \underline{0.7836} & \underline{0.8254} & \underline{0.8107} & \underline{0.8293} & \underline{0.8256} \\
    Mitra & 0.6510 & 0.6012 & 0.6089 & 0.5599 & 0.6304 & 0.5795 & 0.4683 & 0.4173 & 0.8001 & 0.7524 & 0.6471 & 0.6471 & 0.6177 & 0.5650 & 0.4129 & 0.3338 \\
    \underline{TabDPT} & \underline{0.8191} & \underline{0.8157} & \underline{0.8121} & \textbf{\underline{0.8049}} & \textbf{\underline{0.8407}} & \textbf{\underline{0.8299}} & \textbf{\underline{0.7860}} & \textbf{\underline{0.7801}} & \textbf{\underline{0.8609}} & \textbf{\underline{0.8511}} & \textbf{\underline{0.7979}} & \textbf{\underline{0.7979}} & \textbf{\underline{0.8299}} & \textbf{\underline{0.8209}} & \textbf{\underline{0.8416}} & \textbf{\underline{0.8376}} \\
\midrule
\multicolumn{17}{l}{\textit{FineTune - PEFT - Supervised FineTuning}} \\
\cmidrule(lr){1-17}
    TabICL & 0.7957 & 0.7710 & 0.6424 & 0.5686 & 0.6084 & 0.5020 & 0.6003 & 0.5233 & 0.7223 & 0.6507 & 0.6196 & 0.5287 & 0.6596 & 0.5902 & \underline{0.8760} & \underline{0.8502} \\
    OrionBiX & 0.7576 & 0.7102 & 0.6585 & 0.5892 & 0.6327 & 0.5381 & 0.5974 & 0.5164 & 0.7490 & 0.6875 & 0.6278 & 0.5444 & 0.6851 & 0.6177 & 0.6737 & 0.6166 \\
    \textbf{\underline{OrionMSP}} & \textbf{\underline{0.8298}} & \underline{0.8132} & \underline{0.7190} & \underline{0.6597} & \underline{0.6931} & \underline{0.6152} & \underline{0.6753} & \underline{0.6105} & \underline{0.7915} & \underline{0.7423} & \underline{0.6795} & 0.6113 & \underline{0.7477} & \underline{0.6939} & 0.8042 & 0.7663 \\
    Mitra & 0.5244 & 0.4133 & 0.5008 & 0.4013 & 0.5459 & 0.4476 & 0.3189 & 0.2171 & 0.7495 & 0.6497 & 0.5518 & 0.4481 & 0.5106 & 0.4108 & 0.2831 & 0.1874 \\
    \underline{TabDPT} & \underline{0.8269} & \textbf{\underline{0.8182}} & \textbf{\underline{0.8332}} & \textbf{\underline{0.8245}} & \textbf{\underline{0.7686}} & \textbf{\underline{0.7418}} & \textbf{\underline{0.7907}} & \textbf{\underline{0.7798}} & \textbf{\underline{0.8496}} & \textbf{\underline{0.8346}} & \textbf{\underline{0.7821}} & \textbf{\underline{0.7631}} & \textbf{\underline{0.8305}} & \textbf{\underline{0.8201}} & \textbf{\underline{0.8925}} & \textbf{\underline{0.8901}} \\
\bottomrule
\end{tabular}
\end{sidewaystable}

\begin{table}[pt]
  \centering
  \scriptsize
  \caption{Probability calibration metrics across models and tuning strategies on the TALENT, OpenML-CC18, and TabZilla benchmark suites. Values are averaged across datasets within each suite. Models are grouped by adaptation strategy. Formatting: \textbf{\underline{1st place}} ; \underline{2nd place} within each group.}
  \vspace{0.12in}
  \label{tab:calibration}
  \begin{tabular}{l ccc ccc ccc}
    \toprule
Strategy
  & \multicolumn{3}{c}{TALENT}
  & \multicolumn{3}{c}{OpenML-CC18}
  & \multicolumn{3}{c}{TabZilla} \\
\cmidrule(lr){1-1} \cmidrule(lr){2-4} \cmidrule(lr){5-7} \cmidrule(lr){8-10}
Models
  & ECE($\downarrow$) & MCE($\downarrow$) & BS($\downarrow$)
  & ECE($\downarrow$) & MCE($\downarrow$) & BS($\downarrow$)
  & ECE($\downarrow$) & MCE($\downarrow$) & BS($\downarrow$) \\
    \midrule

\multicolumn{10}{l}{\textit{Zero-Shot Inference}} \\
\cmidrule(lr){1-10}
    \underline{TabICL} & \textbf{\underline{0.0219}} & 0.2421 & \underline{0.1533} & 0.0371 & 0.3404 & 0.1267 & \underline{0.0369} & 0.2863 & 0.1301 \\
    OrionBiX & 0.0324 & \underline{0.2245} & 0.1787 & \underline{0.0325} & 0.3230 & 0.1325 & 0.0385 & \textbf{\underline{0.2512}} & 0.2419 \\
    \textbf{\underline{OrionMSP}} & \textbf{\underline{0.0219}} & \textbf{\underline{0.2098}} & 0.1589 & \textbf{\underline{0.0319}} & \underline{0.2902} & \underline{0.1262} & \textbf{\underline{0.0310}} & 0.2805 & \underline{0.1243} \\
    \underline{TabPFN} & \underline{0.0276} & 0.2470 & \textbf{\underline{0.1514}} & 0.0375 & \textbf{\underline{0.2880}} & \textbf{\underline{0.1253}} & 0.0431 & \underline{0.2745} & 0.1283 \\
    Mitra & 0.2238 & 0.3115 & 0.5291 & 0.2138 & 0.2952 & 0.5307 & 0.1946 & 0.2995 & 0.5733 \\
    TabDPT & 0.0308 & 0.2489 & 0.1586 & 0.0443 & 0.3160 & 0.1351 & 0.0435 & 0.2997 & \textbf{\underline{0.1227}} \\
\midrule
\multicolumn{10}{l}{\textit{FineTune - Meta Learning}} \\
\cmidrule(lr){1-10}
    TabICL & 0.0355 & 0.2570 & 0.1671 & \underline{0.0426} & 0.3387 & 0.1346 & 0.2033 & 0.4926 & 0.3732 \\
    OrionBiX & 0.0409 & 0.2693& 0.1841 & 0.0731 & 0.4164 & 0.1607 & 0.0744 & 0.3964 & 0.1416 \\
    \underline{OrionMSP} & 0.0340 & 0.2570 & \underline{0.1656} & 0.0520 & 0.3458 & \underline{0.1606} & \textbf{\underline{0.0365}} & \underline{0.2574} & \textbf{\underline{0.1250}} \\
    \textbf{\underline{TabPFN}} & \textbf{\underline{0.0271}} & \underline{0.2440} & \textbf{\underline{0.1502}} & \textbf{\underline{0.0346}} & \underline{0.2989} & \textbf{\underline{0.1178}} & \underline{0.0388} & 0.2636 & \underline{0.1344} \\
    \underline{Mitra} & \underline{0.0334} & \textbf{\underline{0.1912}} & 0.3479 & 0.0533 & \textbf{\underline{0.2258}} & 0.3932 & 0.0891 & \textbf{\underline{0.2086}} & 0.4239 \\
    TabDPT & 0.0562 & 0.3235 & 0.1791 & 0.0745 & 0.4178 & 0.1620 & 0.0664 & 0.2838 & 0.1591 \\
\midrule
\multicolumn{10}{l}{\textit{FineTune - Supervised FineTuning}} \\
\cmidrule(lr){1-10}
    TabICL & 0.0768 & 0.3219 & 0.2349 & 0.1156 & 0.3610 & 0.3039 & 0.1534 & 0.3280 & 0.4078 \\
    OrionBiX & 0.0753 & 0.3204 & 0.2334 & 0.1131 & 0.3374 & 0.2923 & 0.1581 & 0.3801 & 0.3530 \\
    \underline{OrionMSP} & \underline{0.0471} & \textbf{\underline{0.1886}} & 0.2113 & \underline{0.0550} & 0.3155 & 0.1995 & 0.0773 & 0.3321 & 0.2594 \\
    \textbf{\underline{TabPFN}} & \textbf{\underline{0.0287}} & \underline{0.2509} & \textbf{\underline{0.1580}} & \textbf{\underline{0.0381}} & \underline{0.2957} & \textbf{\underline{0.1246}} & \textbf{\underline{0.0469}} & \textbf{\underline{0.2528}} & \textbf{\underline{0.1454}} \\
    Mitra & 0.1723 & 0.2620 & 0.4677 & 0.1539 & \underline{\textbf{0.2269}} & 0.4784 & 0.2023 & \underline{0.2991} & 0.5254 \\
    \underline{TabDPT} & 0.0548 & 0.3019 & \underline{0.1867} & 0.0497 & 0.3397 & \underline{0.1443} & \underline{0.0713} & 0.3183 & \underline{0.1801} \\
\midrule
\multicolumn{10}{l}{\textit{PEFT FineTune - Meta Learning}} \\
\cmidrule(lr){1-10}
    \underline{TabICL} & 0.0355 & 0.2570 & \underline{0.1671} & \textbf{\underline{0.0426}} & \underline{0.3387} & \textbf{\underline{0.1346}} & 0.2033 & 0.4926 & 0.3732 \\
    OrionBiX & 0.0409 & 0.2693& 0.1841 & 0.0731 & 0.4164 & 0.1607 & 0.0744 & 0.3964 & \underline{0.1416} \\
    \textbf{\underline{OrionMSP}} & \underline{0.0340} & \underline{0.2570} & \underline{\textbf{0.1656}} & \underline{0.0520} & 0.3458 & \underline{0.1606} & \textbf{\underline{0.0365}} & \textbf{\underline{0.2574}} & \textbf{\underline{0.1250}} \\
    \underline{Mitra} & \textbf{\underline{0.0334}} & \textbf{\underline{0.1912}} & 0.3479 & 0.0533 & \textbf{\underline{0.2258}} & 0.3932 & 0.0891 & \underline{0.2086} & 0.4239 \\
    TabDPT & 0.0562 & 0.3235 & 0.1791 & 0.0745 & 0.4178 & 0.1620 & \underline{0.0664} & 0.2838 & 0.1591 \\
\midrule
\multicolumn{10}{l}{\textit{PEFT FineTune - Supervised FineTuning}} \\
\cmidrule(lr){1-10}
    TabICL & 0.0768 & 0.3219 & 0.2349 & 0.1156 & 0.3610 & 0.3039 & 0.1534 & 0.3280 & 0.4078 \\
    OrionBiX & 0.0753 & 0.3204 & 0.2334 & 0.1131 & 0.3374 & 0.2923 & 0.1581 & 0.3801 & 0.3530 \\
    \underline{OrionMSP} & \textbf{\underline{0.0471}} & \textbf{\underline{0.1886}} & \underline{0.2113} & \underline{0.0550} & \underline{0.3155} & \underline{0.1995} & \underline{0.0773} & 0.3321 & \underline{0.2594} \\
    Mitra & 0.1723 & \underline{0.2620} & 0.4677 & 0.1539 & \textbf{\underline{0.2269}} & 0.4784 & 0.2023 & \textbf{\underline{0.2991}} & 0.5254 \\
    \textbf{\underline{TabDPT}} & \underline{0.0548} & 0.3019 & \textbf{\underline{0.1867}} & \textbf{\underline{0.0497}} & 0.3397 & \textbf{\underline{0.1443}} & \textbf{\underline{0.0713}} & \underline{0.3183} & \textbf{\underline{0.1801}} \\
    \bottomrule
  \end{tabular}
\end{table}

\begin{table}[pt]
  \centering
  \scriptsize
  \caption{Fairness metrics across models and tuning strategies on datasets with explicit demographic attributes. Values are averaged across fairness evaluation datasets. Models are grouped by adaptation strategy. Formatting: \textbf{\underline{1st place}} ; \underline{2nd place} within each group.}
  \vspace{0.12in}
  \label{tab:fairness}
  \begin{tabular}{l ccc ccc}
    \toprule
Strategy & \multicolumn{3}{c}{Fairness Metrics} & \multicolumn{2}{c}{Performance Metrics} \\
    \cmidrule(lr){1-1} \cmidrule(lr){2-4} \cmidrule(lr){5-6}
Models & SPD($\downarrow$) & EOD($\downarrow$) & EOpD($\downarrow$) & ACC($\uparrow$) & F1($\uparrow$) \\
    \midrule
\multicolumn{6}{l}{\textit{Zero-Shot Inference}} \\
\cmidrule(lr){1-6}
    TabICL & 0.2900 & 0.3168 & 0.3278 & 0.8743 & 0.8680 \\
    OrionBiX & 0.3328 & 0.2698 & 0.2855 & \textbf{\underline{0.8779}} & \textbf{\underline{0.8727}}  \\
    OrionMSP & 0.3380 & 0.2827 & \textbf{\underline{0.2983}} & \underline{0.8752} & \underline{0.8706} \\
    TabPFN & 0.3070 & 0.3114 & 0.3176 & 0.8708 & 0.8631 \\
    \textbf{\underline{Mitra}} & \textbf{\underline{0.0193}} & \textbf{\underline{0.0590}} & \textbf{\underline{0.0982}} & 0.6902 & 0.5847 \\
    TabDPT & 0.2990 & 0.3197 & 0.3338 & 0.8674 & 0.8582 \\
\midrule
\multicolumn{6}{l}{\textit{FineTune - Meta Learning}} \\
\cmidrule(lr){1-6}
    TabICL & 0.3140 & 0.3100 & 0.3332 & 0.8678 & 0.8603 \\
    \textbf{\underline{OrionBiX}} & \textbf{\underline{0.2021}} & \textbf{\underline{0.1624}} & \textbf{\underline{0.2010}} & \textbf{\underline{0.8743}} & \textbf{\underline{0.8692}} \\
    OrionMSP & 0.3038 & 0.2798 & \underline{0.2907} & 0.8680 & 0.8650 \\
    TabPFN & 0.3070 & 0.3115 & 0.3177 & \underline{0.8733} & \underline{0.8668} \\
    \underline{Mitra} & \underline{0.2586} & \underline{0.2748} & 0.3301 & 0.7967 & 0.7546 \\
    TabDPT & 0.2648 & 0.3180  & 0.3080 &  0.8614 &  0.8503\\
\midrule
\multicolumn{6}{l}{\textit{FineTune - Supervised FineTuning}} \\
\cmidrule(lr){1-6}
    TabICL & \underline{0.0550} & 0.3225 & 0.3319 & 0.4584 & 0.3625 \\
    \underline{OrionBiX} & 0.1128 & \underline{0.1761} & \underline{0.1817} & 0.6180 & 0.5498 \\
    OrionMSP & 0.1227 & 0.2208 & 0.2353 & 0.4753 & 0.4039 \\
    TabPFN & 0.3070 & 0.3115 & 0.3177 & \textbf{\underline{0.8733}} & \textbf{\underline{0.8668}} \\
    \textbf{\underline{Mitra}} & \textbf{\underline{0.0161}} & \textbf{\underline{0.0170}} & \textbf{\underline{0.0555}} & 0.7268 & 0.6277 \\
    TabDPT & 0.3075 & 0.3153 & 0.3284 & \underline{0.8529} & \underline{0.8435} \\
\midrule
\multicolumn{6}{l}{\textit{PEFT FineTune - Meta Learning}} \\
\cmidrule(lr){1-6}
    TabICL & 0.3140 & 0.3092 & 0.3332 & \underline{\textbf{0.8678}} & 0.8602 \\
    \textbf{\underline{OrionBiX}} & 0.3149 & \textbf{\underline{0.2722}}  & \textbf{\underline{0.2875}} & \textbf{\underline{0.8748}} & \textbf{\underline{0.8717}} \\
    OrionMSP & 0.3038 & 0.2798 & \underline{0.2907} & \textbf{\underline{0.867}} & \underline{0.8632} \\
    \underline{Mitra} & \textbf{\underline{0.2595}} & \textbf{\underline{0.2391}} & 0.3154 & 0.8104 & 0.7721 \\
    TabDPT & \underline{0.2778} & 0.2792  & 0.3051 &  0.8651 &  0.8556\\
\midrule
\multicolumn{6}{l}{\textit{PEFT FineTune - Supervised FineTuning}} \\
\cmidrule(lr){1-6}
    TabICL & 0.1572 & 0.4041 & 0.4089 & 0.4881 & 0.4269 \\
    \underline{OrionBiX} & \underline{0.0349} & \underline{0.1564} & \underline{0.2234} & 0.6401 & 0.5759 \\
    OrionMSP & 0.1132 & 0.2140 & 0.2420 & 0.5732 & 0.5080 \\
    \textbf{\underline{Mitra}} & \textbf{\underline{0.0029}} & \textbf{\underline{0.0370}} & \textbf{\underline{0.0740}} & \underline{0.7133} & \underline{0.6028} \\
    TabDPT & 0.2960 & 0.3190 & 0.3342 & \textbf{\underline{0.8595}} & \textbf{\underline{0.8505}} \\
\bottomrule
\end{tabular}
\end{table}
We assess zero-shot and fine-tuned TFMs across performance, calibration, and fairness.
Table~\ref{tab:overall} summarizes results on TALENT, OpenML-CC18, and TabZilla.

Across accuracy and F1, zero-shot inference remains competitive and often outperforms fine-tuned variants, with several models showing reduced performance after SFT.
Across all suites, zero-shot TFMs already achieve strong accuracy and F1, often surpassing or matching their fine-tuned variants. OrionMSP and TabPFN provide the most competitive zero-shot results. Meta-learning offers moderate and architecture-dependent gains, particularly for TabPFN and OrionMSP, but full supervised fine-tuning (SFT) frequently reduces accuracy and F1—most notably for TabICL and OrionBiX. PEFT variants recover part of this lost performance and typically track meta-learning behaviour while retaining their computational efficiency, with TabDPT showing the most stable improvements under PEFT.
Models responses to adaptation vary, reflecting the distinct pretraining processes behind each architecture.

\subsection{Performance Across Dataset Characteristics.}
Table~\ref{tab:comprehensive-analysis} breaks down performance by dataset size, imbalance, and feature dimensionality. 
Fine-tuning is most beneficial on medium-sized datasets (1K–10K samples), where TabPFN and OrionMSP show small but consistent gains. 
On small datasets (<1K), zero-shot inference remains clearly superior, as most models overfit when fine-tuned. On large datasets (>10K), adaptation provides limited benefit beyond zero-shot predictions, with OrionMSP and TabDPT already performing strongly. 
For imbalanced datasets, meta-learning offers the most reliable improvements; SFT continues to degrade several architectures. Wide-feature datasets (>100 features) exhibit the few cases where fine-tuning provides noticeable boosts, such as for TabPFN 
(SFT).
These mixed effects within each regime indicate that dataset heterogeneity—feature correlations, noise, and distributional skew—strongly shapes how models respond to adaptation.

\subsection{Calibration Evaluation :}
As shown in Table~\ref{tab:calibration}, zero-shot TFMs deliver the best overall calibration, with OrionMSP and TabPFN achieving the lowest ECE values across all benchmark suites. Meta-learning largely preserves this reliability while providing modest performance gains. In contrast, SFT substantially worsens calibration for most transformers, increasing both ECE and MCE; TabPFN is the only model that remains consistently well-calibrated after full fine-tuning. PEFT improves stability relative to SFT, but still does not match the calibration quality of zero-shot inference. 
Notably, accuracy and calibration respond differently to adaptation, with accuracy gains rarely translating into improved confidence.

\subsection{Fairness Evaluation : }
Fairness results in Table~\ref{tab:fairness} reveal a clear accuracy–equity trade-off. Mitra achieves the lowest disparity metrics but at significantly reduced predictive performance. Zero-shot and meta-learning generally offer the best balance of accuracy and fairness, with OrionMSP and OrionBiX maintaining competitive accuracy and moderate disparity levels. 
SFT exhibits the largest swings in fairness, especially for TabICL and OrionBiX, whereas TabPFN remains comparatively stable across strategies. PEFT variants show mild fairness improvements in some settings but do not consistently outperform zero-shot or meta-learning.

\subsection{Key Findings : }
Fine-tuning effectiveness is highly model- and data-dependent. TabPFN benefits the most from fine-tuning particularly SFT on medium and wide datasets, while remaining well-calibrated. OrionMSP excels primarily in zero-shot and meta-learning regimes, especially on large or imbalanced datasets. 
TabDPT achieves strong efficiency–performance trade-offs under PEFT. 
In contrast, TabICL consistently degrade under SFT, particularly on small or heterogeneous datasets. 
Overall, fine-tuning yields limited improvements and is most useful in mid-sized or high-dimensional settings. 
Zero-shot inference remains preferable in low-data regimes and when calibration or fairness stability is important.

\textbf{Practical Guidelines : }
\begin{itemize}
    \item \textsc{Prefer zero-shot for small data:} Zero-shot inference is strongest on small datasets (<1K samples) where fine-tuning often overfits.
    \item \textsc{Use fine-tuning selectively:} Apply it mainly on medium-sized (1K–10K) or wide-feature datasets, where TabPFN and TabDPT show consistent benefits.
    \item \textsc{Meta-learning as default:} Meta-learning provides modest but reliable gains and preserves calibration, particularly on imbalanced datasets.
    \item \textsc{Caution with SFT:} Full supervised fine-tuning frequently reduces accuracy and calibration and should be reserved for models tolerant to full updates (e.g., TabPFN).
    \item \textsc{PEFT for efficiency:} PEFT avoids overfitting and recovers most useful improvements, making it especially effective for TabDPT and OrionMSP.
\end{itemize}

Overall, aligning the adaptation strategy with dataset size, feature dimensionality, and calibration requirements is crucial, as zero-shot inference often remains competitive.

\subsection{Limitations : }
Our evaluation focuses on binary and multi-class classification; PEFT is not supported for some models . Sensitive attributes for fairness are manually defined and vary across datasets. For consistentency , results are reported only on the common subset available to all models within each benchmark suite: \texttt{155/181} for TALENT, \texttt{27/36} for TabZilla, and \texttt{63/72} for OpenML-CC18.

\section{Conclusion}

This work demonstrates that tabular foundation models achieve strong zero-shot performance but fine-tuning provides selective benefits that depend heavily on model architecture and dataset characteristics. While meta-learning and PEFT offer moderate gains in specific scenarios, full supervised fine-tuning often degrades performance and calibration. Our guidelines can help practitioners navigate these trade-offs, emphasizing the importance of understanding when adaptation helps versus when zero-shot suffices.
\bibliographystyle{unsrt}
\bibliography{references}
\end{document}